%% file: root.tex
\documentclass[letterpaper, 10 pt, conference]{ieeeconf}  
\pdfoutput=1

\IEEEoverridecommandlockouts                              

\usepackage{graphicx} 
\usepackage{epsfig} 

\usepackage{mathptmx} 
\usepackage{times} 
\usepackage{amsmath} 
\usepackage{amssymb}  
\usepackage{epstopdf}	

\usepackage{todonotes}

\usepackage{cite}
\usepackage[utf8]{inputenc}
\usepackage{mathtools}
\usepackage[ruled,norelsize]{algorithm2e}

\usepackage{subfiles}
\usepackage{standalone}
\usepackage{tikz}
\usetikzlibrary{shapes,arrows}

\usepackage{colortbl}
\usepackage{multirow}

\usepackage{amssymb}
\usepackage[caption=false]{subfig}
\graphicspath{{figures/}{../figures/}}
\usepackage{siunitx}
\DeclareSIUnit\bls{{\percent}bl\per\second}

\usepackage{fancyhdr}
\newcommand{\mytitle}{\textbf{Accepted final version.}
To appear in the proceedings of the \textit{IEEE/RSJ International Conference on Intelligent Robots and Systems 2018}.\\
\copyright 2018 IEEE. Personal use of this material is permitted. Permission from IEEE must be obtained for all other uses, in any current or future media, including reprinting/republishing this material for advertising or promotional purposes, creating new collective works, for resale or redistribution to servers or lists, or reuse of any copyrighted component of this work in other works.}
\DeclareMathOperator*{\argmax}{arg\,max}
\DeclareMathOperator*{\argmin}{arg\,min}
\DeclarePairedDelimiter\abs{\lvert}{\rvert}%

\title{\LARGE \bf
Gait learning for soft microrobots controlled by light fields}

\author{Alexander von Rohr$^{1,2,3}$, Sebastian Trimpe$^{2}$, Alonso Marco$^{2}$, Peer Fischer$^{3,4}$, and Stefano Palagi$^{3,5}$
\thanks{$^{1}$Institute for Electrical Engineering in Medicine, University of Lübeck, 23562 Lübeck, Germany.}%
\thanks{$^{2}$Intelligent Control Systems Group, Max Planck Institute for Intelligent Systems, 70569 Stuttgart, Germany. E-mail: vonrohr@is.mpg.de, trimpe@is.mpg.de, amarco@tuebingen.mpg.de}%
\thanks{$^{3}$Micro, Nano, and Molecular Systems Group, Max Planck Institute for Intelligent Systems, 70569 Stuttgart, Germany. E-mail: fischer@is.mpg.de, palagi@is.mpg.de}%
\thanks{$^{4}$Institut für Physikalische Chemie, University of Stuttgart, 70569 Stuttgart, Germany.}%
\thanks{$^{5}$Max Planck ETH Center for Learning Systems.}
\thanks{S. Palagi is now with the Center for Micro-BioRobotics, Istituto Italiano di Tecnologia, Pisa, Italy. E-mail: stefano.palagi@iit.it}%
\thanks{This work was supported in part by the Max Planck Society, the Cyber Valley Initiative, the Max Planck ETH Center for Learning Systems, and a Max Planck Grassroots grant to S. Palagi and S. Trimpe.}%
}

\begin{document}

\maketitle
\thispagestyle{fancy}
\pagestyle{empty}

\begin{abstract}
Soft microrobots based on photoresponsive materials and controlled by light fields can generate a variety of different gaits.
This inherent flexibility can be exploited to maximize their locomotion performance in a given environment and used to adapt them to changing conditions.
Albeit, because of the lack of accurate locomotion models, and given the intrinsic variability among microrobots, analytical control design is not possible.
Common data-driven approaches, on the other hand, require running prohibitive numbers of experiments and lead to very sample-specific results.
Here we propose a probabilistic learning approach for light-controlled soft microrobots based on Bayesian Optimization~(BO) and Gaussian Processes~(GPs).
The proposed approach results in a learning scheme that is data-efficient, enabling gait optimization with a limited experimental budget, and robust against differences among microrobot samples.
These features are obtained by designing the learning scheme through the comparison of different GP priors and BO settings on a semi-synthetic data set.
The developed learning scheme is validated in microrobot experiments, resulting in a 115\%~improvement in a microrobot's locomotion performance with an experimental budget of only 20 tests.
These encouraging results lead the way toward self-adaptive microrobotic systems based on light-controlled soft microrobots and probabilistic learning control.
\end{abstract}


\section{INTRODUCTION}\label{sec:introduction}

\input{sections/01_introduction}

\section{MICROROBOTIC SYSTEM}\label{sec:microrobot}

\input{sections/02_microrobot}

\section{GAIT LEARNING APPROACH}\label{sec:learning_problem}

\input{sections/03_learning_problem}

\section{LEARNING GAIT FOR 1D LOCOMOTION}\label{sec:learning_1d}

\input{sections/04_learning_1d}

\section{SEMI-SYNTHETIC BENCHMARK}\label{sec:comparison}

\input{sections/05_comparison}

\section{EXPERIMENTAL RESULTS}\label{sec:results}

\input{sections/07_results}


\section{CONCLUSION}\label{sec:conclusion}

\input{sections/08_conclusion}




\section*{ACKNOWLEDGMENT}
The authors thank Dr. Hao Zeng, Dr. Camilla Parmeggiani, Dr. Daniele Martella, and Prof. Diederik S. Wiersma for providing the LCE samples for the microrobots.


\bibliographystyle{bib/IEEEtran}
\bibliography{gait-learning}

\end{document}

%% file: sections/01_introduction.tex
Soft microrobots are mobile robotic devices at the sub-millimeter scale that are made of soft, stimuli-responsive materials, which can act as sensors and/or actuators~\cite{Mourran2016,Huang2016,Breger2015,Zeng2015}.
Indeed, as at the micro scale no robotic components are available, basic robotic functionalities need to be implemented in microrobots by the very materials they are made of.

\begin{figure}[tb]
    \vspace{2mm}
   \centering
   \includegraphics[width = 0.9\linewidth]{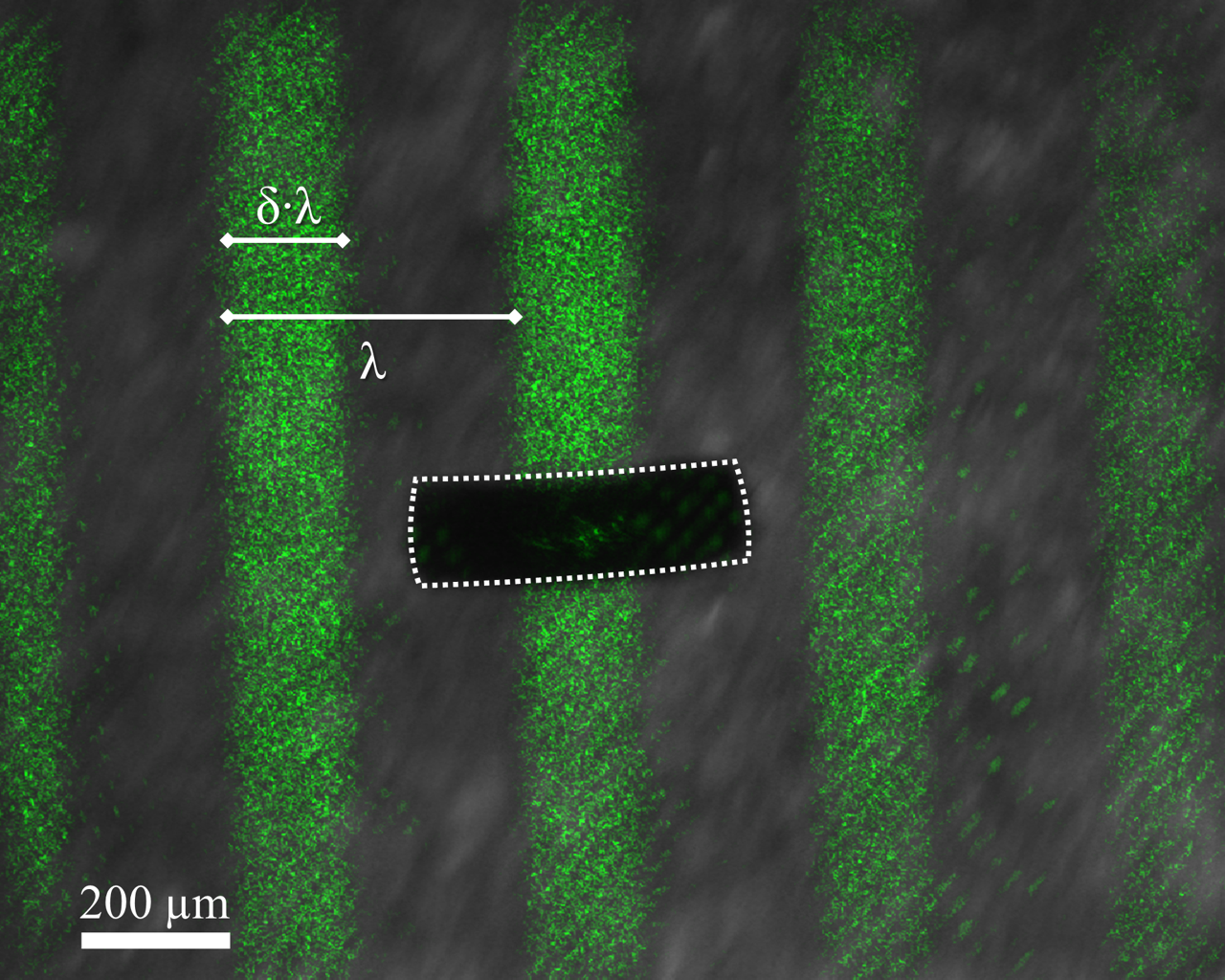}
   \caption{Soft microrobot powered and controlled by structured light field.
    The picture shows the robot (black with dotted outline) and linear light waves with wavelength~$\lambda$ and duty cycle~$\delta$; in this example, $\lambda=300$~pixels (about \SI{390}{\micro\meter}) and $\delta=\SI{40}{\percent}$.
    The pattern will move to the right in the next time step.}
   \label{fig:light_pattern}
\end{figure}

Recently, soft microrobots consisting of photo-responsive liquid-crystal elastomers (LCEs) have been presented~\cite{Palagi2016Nat}.
The material-based actuation of these soft continuum robots is powered and controlled by structured light fields (see Fig.~\ref{fig:light_pattern}), resulting in microrobots with many internal degrees of freedom (DOFs)~\cite{Palagi2016MARSS}.
We define a gait, in the context of the microrobotic system, as any pattern of periodic deformation that results in locomotion.
The deformation and, thus, the gait of these soft microrobots can be flexibly controlled and tuned by the driving light field, enabling the adaptation of the gait to different environments~\cite{Palagi2017MARSS}.
Nonetheless, finding good locomotion performance by hand-tuning the light field parameters is extremely inefficient and time-consuming, especially considering that: (i) often there is no accurate model of the microrobot locomotion; (ii) each microrobot is to some extent different because of their manual fabrication process; and (iii) the response of the microrobot changes over time as the material slowly degrades.
Therefore, learning effective gaits from data is an attractive approach.

In this paper, we propose an automatic learning procedure to find the optimal gait for light-driven soft microrobots in a given environment.
The optimal controller's parameters, i.e.\ the parameters of the driving light field, are progressively evaluated by Bayesian Optimization (BO).
Learning in this scenario presents a few peculiar challenges.
First, the developed learning scheme should be highly data efficient, making use of prior information and leveraging empirical data to approach optimal locomotion performance in a limited number of experiments.
At the same time, it should be robust against differences among microrobots, performing well independently of the specific microrobot used.

This work represents the first attempt to learn a controller from data for sub-millimeter robots.
First, we define a method to efficiently evaluate the locomotion speed from the microrobot's tracking images.
We then report a comparison of a number of BO settings (i.e.\ acquisition functions, kernels, hyperparameters) on a semi-synthetic dataset, which is based on actual data from microrobot experiments and augmented to model inter-microrobot variability.
Finally, we demonstrate successful learning in real-robot experiments, with a locomotion performance improvement of \SI{115}{\percent} with respect to an informed initial guess.
The experimental validation was performed using a previously untested microrobot.

The paper is organized as follows.
The microrobotic system is described in Sec.~\ref{sec:microrobot}, and the proposed gait learning approach is introduced in Sec.~\ref{sec:learning_problem}.
Section~\ref{sec:learning_1d} presents the microrobotic controller and the extraction of prior information from previously published experiments.
The benchmarking of a number of BO settings on a semi-synthetic dataset is described in Sec.~\ref{sec:comparison}, whereas Sec.~\ref{sec:results} reports successful learning in microrobot experiments.

%% file: sections/02_microrobot.tex
In this work, we learn optimal gaits for the soft microrobots developed in~\cite{Palagi2016Nat}.
These consist of monolithic LCEs that actively and continuously deform in response to light.
As the deformation occurs locally, desired motion patterns can be obtained by exciting the microrobots with proper light fields.
This approach results in microrobots having many internal DOFs~\cite{Palagi2016MARSS}.
In other words, the light-based control allows these sub-millimeter monolithic structures to behave as if they contained many independent, wirelessly controllable actuators.

The driving light fields are generated by modulating the intensity of a laser beam and projected onto the microrobots workspace by a microscope objective~\cite{Palagi2016Nat}.
The microrobot is observed through the same objective by a camera that images the workspace at 10 frames per second and at a resolution of 1280$\times$1024~pixel ($\SI{1}{pixel}=\SI{1.29}{\micro\meter}$).
The laser beam (532~nm -- Verdi~G10, Coherent) is modulated in space and time by means of a computer-controlled digital micromirror device (DMD) module (V-7000, ViaLUX).
The DMD consists of an array of 1024$\times$768 micromirrors, each representing a pixel in the projected light field.
The micromirrors modulate the light intensity in a binary fashion, yet gray-scale levels control can be achieved by high-frequency pulse-width modulation (PWM).
Therefore, the microrobotic system has, in principle, almost~10$^6$~control parameters that can each assume 2$^8$~values.
This represents a huge parameter space for gait optimization.
Learning controller parameters directly in such a high-dimensional parameter space would require a prohibitive amount of experimental data.
Therefore, at this stage, we propose learning gaits by employing an effective controller structure as in~\cite{Palagi2016Nat} and tuning its critical free parameters with BO.

%% file: sections/03_learning_problem.tex
\begin{figure}[ht]
    \vspace{2mm}
    \centering
    \includegraphics[width=0.90\linewidth]{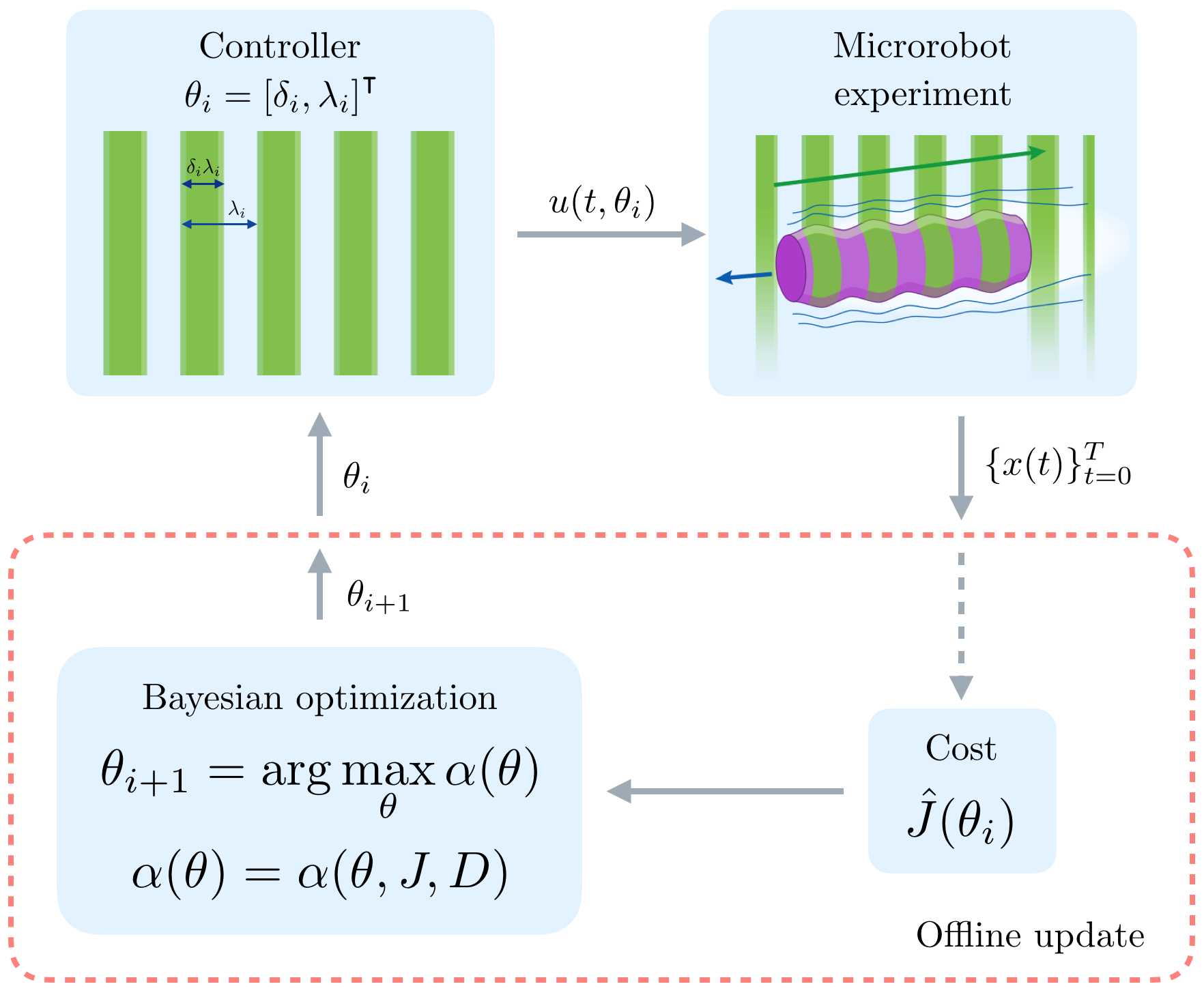}
    \caption{Learning controller parameter from data.
    In each iteration $i$, the controller generates a different control input $u$ for the microrobot. The control input in our case is the dynamic light field (symbolized by the green stripes) parameterized by $\theta_i$.
    The microrobot's state trajectory $\{x(t)\}^T_{t=0}$ during the experiment is tracked and its performance evaluated by the cost function.
    The resulting cost $\hat{J}$ is given back to the optimizer and added to the dataset $D$.
    Based on the GP posterior $J|D$, the acquisition function $\alpha$ is maximized and a new set of parameters is obtained for the next experiment.}
    \label{fig:learning_loop}
\end{figure}

In this section, we formulate the problem of gait learning for light-driven soft microrobots as a cost-minimization problem over a suitably parametrized open-loop controller.
The gait learning algorithm is schematically represented in Fig.~\ref{fig:learning_loop}.
Following the approach in \cite{Palagi2016Nat}, we use periodic light patterns with only few open parameters.
The optimization then takes place in the parameter space of the light pattern.
We also provide here a brief overview of BO with Gaussian Processes (GPs)~\cite{rasmussen2006gpml}, which represents the core of he gait learning method presented herein.

BO with GPs has successfully been used, for example, for gait learning with bipedal walkers~\cite{calandra2016bayesian}, quadrupedal robots~\cite{lizotte2007automatic}, and in cm-scale hexapodal robots~\cite{yang2018learning}.
It has also been proposed for automatic feedback controller tuning~\cite{Marco2016,BeScKr16,marco2017virtual}.
A major strength of GPs is that they allow one to include existing information about the system in the form of a probabilistic prior.
This reduces the size of the data set needed to learn a good approximation of the cost function, while at the same time retaining the flexibility of a non-parametric model.

\subsection{Controller Learning}

We formulate the problem of gait learning in soft microrobots as a parametric controller tuning problem.
This work builds on the automatic controller tuning methods proposed in~\cite{Marco2016}, where an unknown controller cost function was learned in a data-efficient way using BO with GPs.

The microrobot is considered an uncertain, non-linear dynamic system.
The dynamics of a specific microrobot in a given environment are unknown and are evaluated by carrying out experiments.
The experimental observations are affected by uncertainty and noise.
The general state space model of the microrobot is
\begin{equation} \label{eq:state_space}
  \begin{aligned}
    \dot{x}(t) &= g(x(t), u(t), w(t))
  \end{aligned}
\end{equation}
where $x(t) \in \mathbb{R}^{d_x}$ is the system's state (which, we assume, can be fully observed or estimated), $u(t) \in \mathbb{R}^{d_u}$ the input, and $w(t)$ is the process noise.

The input signal $u(t)$ is generated by a controller $C$, which is a function of a set of bounded parameters $\theta \in \Theta \subset \mathbb{R}^{d_c}$ and of the desired movement reference $r(t)$, namely,
\begin{equation} \label{eq:control_output}
  u(t) = C(\theta,r(t),t).
\end{equation}

For gait learning, we need to evaluate how each input sequence influences the unknown system dynamics~\eqref{eq:state_space}.
For this, a cost function $J: \Theta \rightarrow \mathbb{R}$ is used to map the controller parameters $\theta$ to a scalar cost value given the experimental data.
Learning an optimal gait for the given objective corresponds to finding an optimal $\theta^*$ that minimizes $J$:
\begin{equation}
  \theta^* = \argmin_{\theta \in \Theta} J(\theta).
  \label{eq:gait_cost}
\end{equation}

Due to the process noise~$w(t)$ in~\eqref{eq:state_space}, however, we never directly observe $J(\theta)$ but a noisy version of it,~$\hat{J}(\theta) = J(\theta) + \varepsilon$, where $\varepsilon$ is assumed zero-mean Gaussian noise with variance~$\sigma_n^2$.

\subsection{Gaussian Process}
\label{sec:gp}

To account for uncertainty in measurements and state estimation, as well as to formulate prior knowledge about the cost function, if such knowledge is available, the cost function $J(\theta)$ is modeled as a GP.
A GP can be viewed as a distribution over a space of functions. It is a non-parametric model, defined by its mean and covariance functions, $\mu(\theta)$ and $\text{cov}(J(\theta),J(\theta')) = k(\theta,\theta')$, where $k$ is the kernel function
\begin{equation}
  J(\theta) \sim GP(\mu(\theta), k(\theta,\theta')).
\end{equation}

Given a set of $N$ samples, $D = \{\theta_i,\hat{J}(\theta_i)\}_{i=1}^N$, the posterior at $\theta$ is
\begin{equation}\label{eq:posterior}
  \begin{aligned}
    J_{\text{post}}(\theta)|D &\sim N(\mu_{\text{post}}(\theta),\sigma_{\text{post}}^2(\theta)), \text{ where} \\
    \mu_{\text{post}}(\theta) &=  \mu(\theta) + k^\intercal(\theta)K^{-1}y\\
    \sigma_{\text{post}}^2(\theta) &= k(\theta,\theta) - k^\intercal(\theta)K^{-1}k(\theta)\\
  \end{aligned}
\end{equation}
where $\;k(\theta),y \in \mathbb{R}^N$ with $y_i=\hat{J}(\theta_i)-\mu(\theta_i)$, $k(\theta)_i=k(\theta,\theta_i)$, and $K \in \mathbb{R}^{N \times N}$ is the covariance matrix of the training points plus the noise variance $ K_{l,m} = k(\theta_l,\theta_m) + \sigma_n^2\delta_{l,m},\;l,m \in \left\lbrace 1,2,\dots,N \right\rbrace $, where $\delta_{l,m}$ is the Kronecker delta.

While the prior mean is often assumed to be constant, the choice of the kernel function encodes our assumptions and prior knowledge about the cost function.
The kernel function usually depends on a set of hyperparameters.
Typical hyperparameters are the length scales~$l_c$ for each dimension of the parameter space~$\mathbb{R}^{d_c}$ and the signal variance~$\sigma^2_f$.
The kernel's length scales are related to the rate of variation of the underlying function, where short length scales correspond to quickly-varying functions and long length scales to slow-varying ones.
The signal variance describes the width of the distribution and relates to the magnitude of variation around the mean.

In principle, kernel hyperparameters may be obtained from problem knowledge, or estimated from data, e.g., by maximizing the marginal likelihood of the observation with respect to the hyperparameters.  Here, we follow a hybrid approach by setting a prior belief over the hyperparameters (\emph{hyperprior}), and using maximum a posteriori (MAP) estimation for the likelihood, i.e., maximizing marginal likelihood times hyperprior.
We can thus optimize hyperparameters of the GP after each experimental evaluation.
Encoding knowledge in the form of hyperpriors facilitates hyperparameter learning when the data set is small, as is the case for the problem herein, while still retaining flexibility by adjusting hyperparameters based on observed data.
In this work the hyperparameters are estimated from previous experimental results, as reported in Sec.~\ref{sec:hyper}.

\subsection{Bayesian Optimization}
\label{sec:bo}

As the system dynamics are unknown, the gait performance, captured by the cost function \eqref{eq:gait_cost}, can only be optimized by \emph{sampling}, i.e.\ by conducting experiments on the physical system.
BO is a derivative-free global optimization algorithm, which is particularly suited for problems with costly and uncertain function evaluations~\cite{shahriari2016taking}.

Given the current belief about the cost function (i.e.\ the posterior GP~\eqref{eq:posterior}), BO suggests the parameters $\theta_{i+1}$ for the next evaluation by maximizing an acquisition function $\alpha$,
\begin{equation}\label{eq:acquisition_maximization}
  \theta_{i+1} = \argmax_{\theta \in \Theta} \alpha(\theta).
\end{equation}.

The acquisition function is formulated to capture the utility of knowing the outcome of the experiment at the suggested point, for example, the probability of improvement or some expected information gain.
The concrete acquisition functions used for the gait learning problem are described in Sec.~\ref{sec:acquisition_function}.
After evaluating $\hat{J}(\theta_{i+1})$, the new data point is added to $D$, the GP posterior~\eqref{eq:posterior} updated, and BO enters the next iteration.

%% file: sections/04_learning_1d.tex
In the previous section, we presented gait learning for microrobots as an optimization problem, where parameters of an open-loop controller are optimized from experimental trials via BO.
In this section, we consider a concrete instance of a gait learning problem, namely learning the one-dimensional (1D) movement of cylindrical microrobots submerged in a liquid (see Fig.~\ref{fig:microrobot_picture}).
In particular, the microrobots are placed on the bottom of a Petri dish (which is coated with Polydimethylsiloxane (PDMS) to avoid adhesion) and submerged in silicone oil, as in~\cite{Palagi2016MARSS}.
The locomotion of the microrobots occurs as a result of their periodic deformation, which is powered and controlled by the projected light fields.

This section continues by describing the light patterns used to control the microrobot and their defining parameters.
We then define the movement objective and the corresponding cost function used to learn the 1D gait.
Lastly, we estimate the GP hyperparameters (cf.\ Sec.~\ref{sec:gp}).

\subsection{Light Pattern}\label{sec:light_pattern}
The gait objective for this learning task is locomotion along one dimension, for which linear traveling waves of deformation are generated in the microrobot's body.
The deformation is controlled by the light field, which consists of a linear stripe pattern (see Fig.~\ref{fig:light_pattern}).
The controller $C(\theta)$ generates the light field with intensity~$I$, defined as a rectangular wave in the 2D workspace~$(x_s,y_s)$ of the microrobot.
This light field represents the controller output $u$ from~\eqref{eq:control_output}.
As the microrobot is aligned along the $x_s$-axis, the light pattern can be expressed as
\begin{equation}
  I(x_s,y_s,t)=
  \begin{cases}
    1,      & \text{if}\ \ \abs{\mathrm{mod}(x_s/\lambda-ft)}{} \leq \delta/2 \\
    0,      & \text{otherwise.}
  \end{cases}
\end{equation}
The pattern is parametrized by the duty cycle~$\delta$, the spatial wavelength~$\lambda$, and the time domain frequency~$f$.
In the reported experiment, we set the frequency~$f$ to~\SI{1}{\hertz}, whereas the parameters
$\theta = [\lambda, \, \delta]^\intercal$
are to be optimized.

\subsection{Linear Movement}

Swimming at the microscale is governed by the Stokes equations, where inertial forces play a negligible role with respect to viscous forces.
In other words, a microrobot generating a constant propulsive force will swim at a constant speed~$V_m$, and its acceleration to this speed will be instantaneous.
Therefore, the acceleration of the microrobot is neglected and its state space consists of two dimensions, namely the microrobot position $x_m$ and its speed $v_m=d{x_m}/dt$.
However, because of the control scheme described in Sec.~\ref{sec:light_pattern}, the position of the microrobot's center of mass will have some oscillation at the frequency~$f$ of the traveling wave deformation (see Fig. \ref{fig:movement_fit}).
We are interested in optimizing the linear component of the propulsion, regardless of the oscillations caused by the exciting light field.
Therefore we consider only the microrobots constant speed $V_m$.
As the cost in~\eqref{eq:gait_cost}, we use the deviation of this constant speed component from the desired, $J = \abs{V_m^* - V_m}$.
The estimation of $V_m$ from the microrobot's position tracking is explained in the next section.

\subsection{Velocity Estimation}
\label{sec:estimation}
\begin{figure}[tb]
    \vspace{2mm}
   \centering
   \subfloat[Capture of a microrobot\label{fig:microrobot_picture}]{%
        \includegraphics[width=0.9\linewidth]{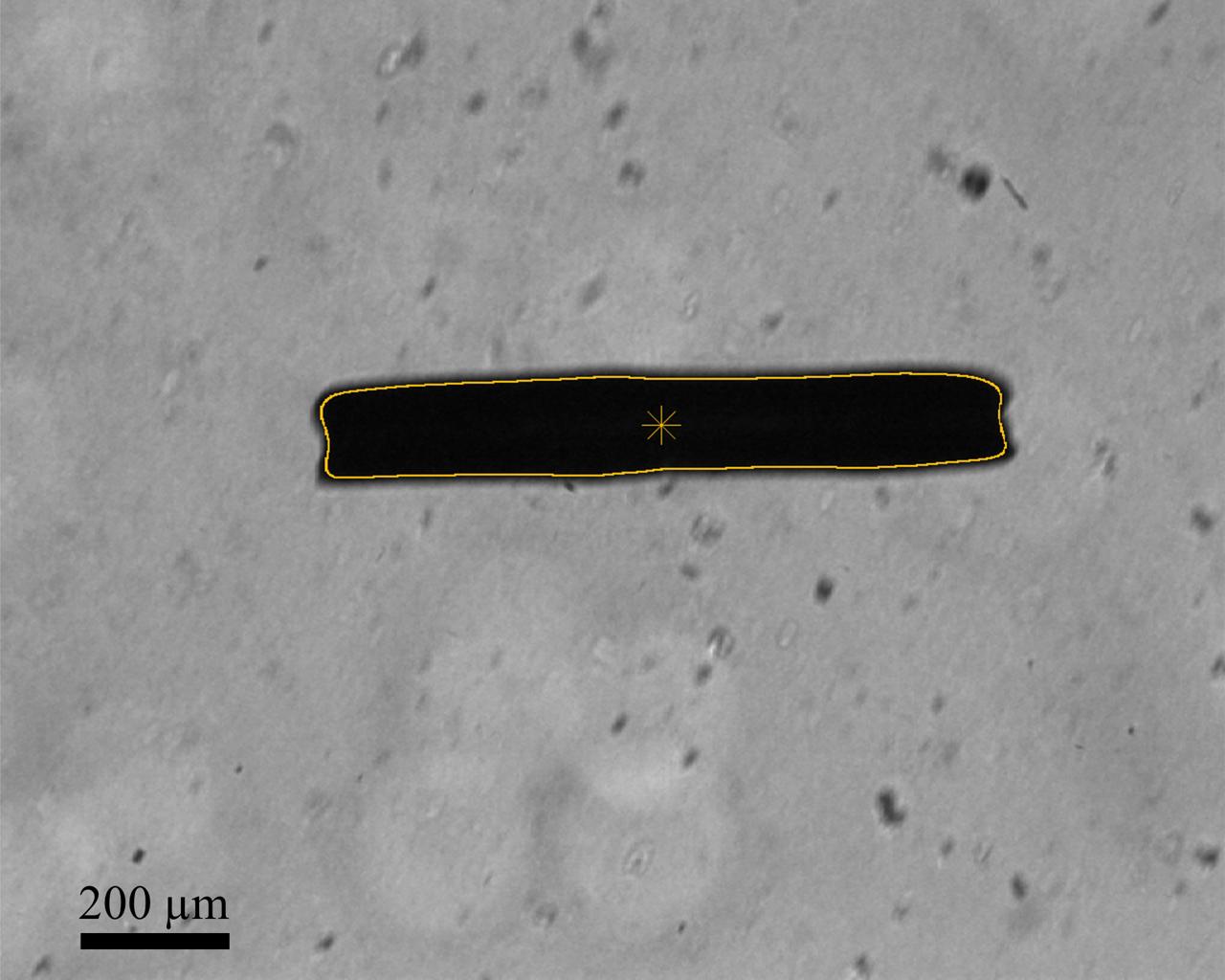}}
        \\
   \subfloat[Average speed estimation\label{fig:movement_fit}]{%
        \includegraphics[width=0.9\linewidth]{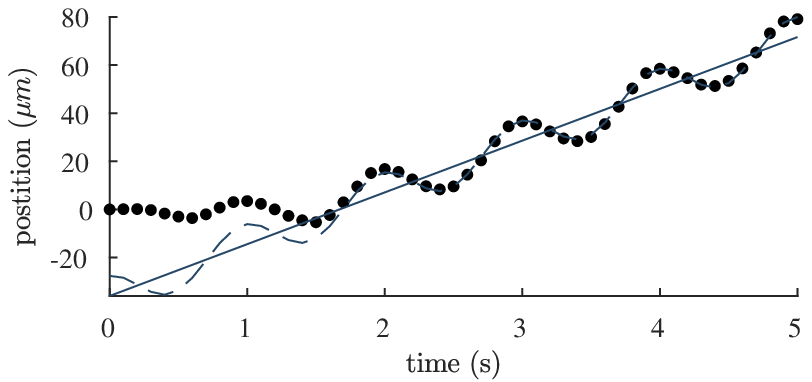}}
   \caption{Velocity estimation. (a) The microrobot during operation from above as seen from the camera. The position is determined by the center of mass (yellow star) of the area enclosed by the estimated edges (yellow line).
  (b) The position of the microrobot $x_m$ during the experiment is shown by the black dots. The dashed blue line shows the fitted model. The solid blue line represents the linear part of the movement. Its slope is the average speed $V_m$.}
   \label{fig:microrobot_movement}
\end{figure}

We estimate the velocity based on the experimental setup described in Sec.~\ref{sec:microrobot}.
The position~$x_m(t)$ of the microrobot is calculated by binarizing the acquired images and calculating the center of mass of the area corresponding to the microrobot (see Fig.~\ref{fig:microrobot_picture}).

One possible approach to obtain the underlying constant speed $V_m$ from the observed state trajectory $\{x(t)\}^T_{t=0} := \{x_m(t)\}^T_{t=0} $ consists in estimating the instantaneous velocity $v_m(t)$, by differentiating the position $x_m(t)$, and then averaging over an integer number of periods $\tau=1/f$.
As derivatives are highly susceptible to noise and the sampling frequency is close to the oscillation frequency, we chose to use a different approach.
We estimate the average speed $V_m$ \emph{directly} from the position by fitting the phenomenological model
\begin{equation}
  x'_m(t) = (V_m t + b)\ +\ a \sin(2 \pi f t + \phi).
  \label{eq:movement_model}
\end{equation}

This model captures the two types of movement of the soft microrobots, namely the linear movement at constant speed~$V_m$ and the oscillations at frequency~$f$.
The oscillation term is defined by the known frequency~$f$, and the unknown amplitude~$a$ and phase shift~$\phi$.
The linear term includes the offset~$b$, which accounts for changes in the $x_s$ position that occur during the transient (cf.\ Fig.~\ref{fig:movement_fit}).
In previous experiment we observed the presence of a transient, which is not due to acceleration but to the initial response time of the material with respect to the start of the light projection.
As we have no phenomenological model of the transient, but previous experiments have shown that it decayed after about two seconds, we only fit the model starting at $t=\SI{2}{\second}$.
Fig.~\ref{fig:movement_fit} shows the measurement for one experiment as well as the fitted movement model.
The constant speed $V_m$ is the slope of the linear part represented by the solid line.

The speed estimation procedure also allows for quantifying the uncertainty of the velocity estimate, which we did not use, however, for the experiments herein.

\subsection{Hyperparameter Estimation}
\label{sec:hyper}
Estimates of the hyperparameters are obtained from the experimental data reported in~\cite{Palagi2016MARSS}.
For this, the raw data from these experiments are re-processed to estimate the speed according to the model~\eqref{eq:movement_model}.
These data are then used to estimate the boundaries of the parameter space and the prior for the cost function hyperparameters, namely, the constant mean~$\mu$, the signal variance~$\sigma_f^2$ for the kernel, and the noise variance $\sigma_n^2$.
The kernel length scales are more difficult to estimate with the given data and are instead set to one fourth of the total range for each parameter.

To account for variability in the microrobot size, the estimated speed is normalized on the bodylength, as common in microrobotics, and is expressed in bodylength percent per second (\si{\bls}).
The fastest speed observed in these experiments, according to the velocity estimation described in Sec. \ref{sec:estimation} is \SI{4.34}{\bls}, obtained with a duty cycle $\delta=\SI{30}{\percent}$ and a wavelength $\lambda=\SI{645}{\micro\meter}$.
From the measurements, we estimate the signal standard deviation~$\sigma_f$ to be \SI{2}{\bls} and the noise standard deviation~$\sigma_n$ to be \SI{0.1}{\bls}.
Physically reasonable ranges for the parameter space are a duty cycle $\delta$ comprised between \SI{20}{\percent} and \SI{50}{\percent} and a wavelength ranging from $200$ to $800$~pixels in the camera frame (corresponding to $258$ to \SI{1032}{\micro\meter}).

As prior distributions for each hyperparameter, we use Gaussian distributions with the corresponding estimated value as mean and one fourth of the estimate as standard deviation.

%% file: sections/05_comparison.tex
For BO to work well in practice, we need to appropriately choose some settings, in particular, the kernel and the acquisition function.
Appropriate choices for these should allow for generalization to different microrobots, while retaining data-efficiency.
To determine a good setup in a systematic way, we consider a semi-synthetic cost function and benchmark different settings.

In this section, we first describe how the semi-synthetic cost function is generated, then we motivate the considered design choices for the GP and BO, and finally we present the benchmarking results.

\subsection{Creating a Semi-Synthetic Cost Function}

As the systems dynamics~\eqref{eq:state_space} are unknown, we cannot properly benchmark the aforementioned BO settings on a pure simulation.
Furthermore, a good performance on a fully synthetic optimization test function will not necessarily translate to real-world performance gains.
We thus propose to sample from the real system on a coarse grid, and then interpolate the data at unobserved locations to create a semi-synthetic cost function, which has similar properties to the real one, while being cheaper to evaluate.
To collect the necessary data, we defined a grid on the parameter space and used BO with a standard kernel to sample from this grid.
Specifically, we collected 56 cost function evaluations by running BO with a squared exponential (SE) kernel and parameters defined in~\ref{sec:hyper}.
The observations were conducted with the microrobotic system described in Sec.~\ref{sec:microrobot} and with the experimental conditions reported in Sec.~\ref{sec:learning_1d}.
These are the same conditions as used in~\cite{Palagi2016MARSS}, from which we estimated the prior information.
We use a different microrobot which results in a slightly different locomotion behavior.

\begin{figure}[thpb]
   \centering
   \includegraphics[width=0.90\linewidth]{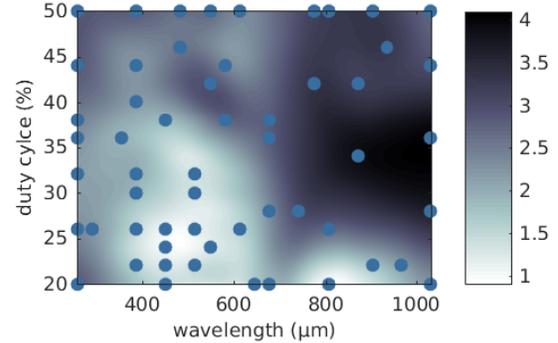}
   \caption{Contour plot of a random sample of the semi-synthetic cost function. Dots are locations of measurements from a real microrobot.
   }
   \label{fig:benchmark}
\end{figure}

For each BO run, we generate different semi-synthetic cost functions by considering the intrinsic noise of the measurements and re-sampling each data point.
This assures that the cost function, including the location of the optimum, is different for each run, while retaining the overall `shape' to some extent.
As not all points on the grid were actually evaluated in experiments, the missing data points were generated by linear interpolation.
The resulting data was smoothed with a 3$\times$3 spatial mean filter.
To obtain the final continuous cost function, the discrete data set was further interpolated using cubic splines.
Figure~\ref{fig:benchmark} shows one sample from the possible benchmark functions.

\subsection{Gaussian Process Priors}
\label{sec:gp_design}

In this section, we describe the choice of the prior mean, signal variance, and kernel function, to model the semi-synthetic cost function.

We distinguish between two kinds of priors: \emph{optimistic} and \emph{pessimistic}. An optimistic prior assumes good performance (i.e.\ low cost values) at unobserved locations, while a pessimistic prior assumes the opposite. Since we terminate BO after a small number of experiments, the choice of the prior has an impact on the search: An optimistic prior encourages an extensive exploration, while a pessimistic prior explores slower.

\subsubsection{Mean}

As prior mean function we use a constant non-zero mean.
For a cost function that is close to zero at its minimum, a mean of zero would be optimistic.
If the real cost function has in fact a higher mean, a zero mean prior would make the optimizer believe that it has initially evaluated bad regions of the parameter space and therefore keep exploring new regions.
This leads to an undesirably high amount of exploration at the corners and edges of the feasible parameter space.

With a more pessimistic prior mean (e.g.\ at approximately the standard deviation of the signal, $\mu \approx \sigma_f$) the belief is that most controllers perform badly.
We found that this yields a much better exploration-exploitation trade-off.
All settings presented herein use a pessimistic mean of $\mu = \sigma_f$ as prior mean.

\subsubsection{Signal Variance}

With a pessimistic mean $\mu>0$, we can readjust the estimated value for the signal variance $\sigma_f^2$ from the initial estimate used to find $\mu$ such that the lowest possible cost, which is $0$ by definition, is either inside or outside the \SI{95}{\percent} confidence interval of the prior.
We call the first case an optimistic setting and the second a pessimistic one.
During the benchmark we compared an optimistic signal variance, such that $\sigma_{f1}>\mu/2$, to a pessimistic one, with $\sigma_{f2}<\mu/2$, to evaluate the effect of this setting on the BO performance.

\subsubsection{Kernel}

The kernel is the main object in the GP that encodes prior knowledge.
We compare four standard kernels and one composite kernel.

The SE kernel $k_{SE}(\theta,\theta')$ is one of the most frequently used kernels for GP regression~\cite{rasmussen2006gpml}.
We use SE with automatic relevance determination (ARD), i.e.\ we have different length scales for each dimension.

Another common option is the Rational Quadratic (RQ) kernel $k_{RQ}(\theta,\theta')$ with ARD~\cite{rasmussen2006gpml}.
This kernel can be seen as a weighted sum over SE kernels of all lengthscales.

The SE and RQ kernels both assume very smooth functions~\cite{rasmussen2006gpml}. While the smoothness assumption of the SE and RQ kernels is too restrictive for most practical optimization problems (e.g.\ physical systems), the Matern kernel is known to be a much better choice~\cite{rasmussen2006gpml,snoek2012practical}.
We compare two versions of the $\nu$-Matern ($\nu$M) kernel with ARD,
$k_{M32}(\theta,\theta')$ (where $\nu = 3/2$) and $k_{M52}(\theta,\theta')$ (where $\nu = 5/2$). The $\nu$ property of the Matern kernel determines the smoothness of modeled functions, where a higher $\nu$ yields smoother functions.

We also adopt a composite kernel, which we term $2Mat$, having one component that captures changes of the cost function at a short length scale, and one smoother component for variations at a longer scale:
\begin{equation}
  k_{2Mat}(\theta,\theta') = k_{M52}(\theta,\theta') + k_{M32}(\theta,\theta')
\end{equation}
The sum of two kernels is again a valid kernel, \cite{rasmussen2006gpml}. Combining simple kernels, e.g., with different length scales, is a powerful way to build more expressive models.

\subsection{Acquisition Functions}
\label{sec:acquisition_function}

As mentioned in Sec.~\ref{sec:bo}, the adopted acquisition function determines the behavior and performance of BO.
An acquisition function evaluates the utility of a candidate for sampling.
Here a candidate is a parameter $\theta$ for the controller $C$ in the feasible parameter space.
BO then chooses as sampling location the candidate with the maximum utility.
Here we test the following acquisition functions.

\emph{Probability of Improvement} (PI) \cite{lizotte2008practical} is the probability that the candidates cost value is below some threshold $\gamma \mu^*$.
Here $\mu^*$ is the smallest cost as predicted by the posterior mean and $\gamma$ a constant factor used to trade off exploration vs. exploitation.
We set $\gamma$ to an optimistic value of $0.9$, expecting that the posterior is pessimistic, as its prior is a pessimistic mean.

\emph{Expected Improvement} (EI) \cite{jones1998efficient} is the expected value of improvement over $\gamma \mu^*$ with an optimistic $\gamma$ of $0.9$.
In contrast to PI, EI takes into account not only the probability of an improvement in the cost function at a candidate position, but also the expected magnitude of such improvement.

\emph{Entropy Search} (ES) \cite{hennig2012entropy} evaluates the expected information gain (change in entropy) about the location of the minimum at the candidate.
As such, ES is not directly concerned about sampling at the location of the minimum, but about samples that maximize the information gain about the minimum location.

\subsection{Results}

\begin{table*}[tb]
\vspace{2mm}

\caption[]{Benchmarking on a semi-synthetic function: Overall best in bold, highlighted benchmarks are detailed in Fig.~\ref{fig:comp_a};\\
median (95-percentile) over 200 runs of the normalized regret~$R$~(\%) after 20 iterations.}
	\label{tab:results}
	\begin{center}
		\begin{tabular}{c||c|c|c|c|c|c|c}

			\multirow{2}{*}{} & \multicolumn{2}{c|}{EI $\sigma_{f1}$}    & \multicolumn{2}{c|}{EI $\sigma_{f2}$}    & \multicolumn{2}{c|}{PI}                   & ES                  \\ \cline{2-8}
			        & fixed Hyp.                              & learned Hyp.                            & fixed Hyp.                      & learned Hyp.                   & fixed Hyp.  & learned Hyp. & fixed Hyp.                     \\ \hline \hline
			SE     & \cellcolor{gray!60} 11.8 (54.3)         & 10.7 (43.8)                             & 11.4 (53.2)                     & 14.9 (59.7)                    & 9.9 (47.5)  & 13.8 (49.4)  & 11.7 (51.1)                    \\ \hline
			RQ      & 5.5 (35.3)                              & 4.7 (27.6)                              & 7.5 (37.8)                      & 7.3 (46.2)                     & 7.3 (42.8)  & 10.1 (66.1)  & \cellcolor{gray!60} 6.0 (46.1) \\ \hline
			M32     & 5.8 (37.7)                              & 3.3 (33.2)                              & 4.1 (33.6)                      & 4.4 (36.2)                     & 5.8 (51.2)  & 7.4 (66.2)   & 7.5 (43.9)                     \\ \hline
			M52     & 3.5 (29.2)                              & 4.2 (33.9)                              & \cellcolor{gray!60} 20.0 (62.4) & \cellcolor{gray!60} 4.5 (37.6) & 13.5 (71.8) & 8.0 (53.3)   & 7.9 (47.7)                     \\ \hline
			2Mat & \cellcolor{gray!60} \textbf{3.4 (23.0)} & \cellcolor{gray!60} \textbf{3.5 (26.7)} & 5.4 (43.8)                      & 7.0 (38.6)                     & 6.8 (41.1)  & 5.6 (33.4)   & -                              \\
		\end{tabular}
	\end{center}
\end{table*}

The results for BO on the semi-synthetic cost functions serve to evaluate the relative performances of different design choices.
While we do not necessarily expect to achieve the exact same absolute performance on a real microrobot, the comparison gives a good indicator on which setting to prefer, when learning microrobot gaits.

All experiments start at the same location, i.e.\ the initial controller found in Sec.~\ref{sec:hyper} ($\delta=\SI{30}{\percent}$, $\lambda=\SI{645}{\micro\meter}$).
For EI we tried both an optimistic and a pessimistic signal variance, respectively~$\sigma_{f1}^2$ and~$\sigma_{f2}^2$ (see above), whereas PI and ES were tested only with the optimistic signal variance~$\sigma_{f1}^2$.

Each design choice of the BO was run on 200 different semi-synthetic cost function.
Each BO run consisted of 20 iterations on one cost function, representing 20 experimental observations on the same system (the experimental budget).
For each iteration, we calculated the normalized regret $R$ after each iteration defined as
\begin{equation}\label{eq:norm_cost}
R=\frac{ J(\theta^*) - J(\theta^{\text{opt}})} {J(\theta^{\text{opt}})}
\end{equation}
where $\theta^*$ is the best parameter set as predicted by the GPs posterior mean, and $\theta^{\text{opt}}$ is the parameter set at the true optimum.

\begin{figure}[tb]
 \centering
 \subfloat[Median regret over BO iteration\label{fig:comp_a}]{%
      \includegraphics[width=0.90\linewidth]{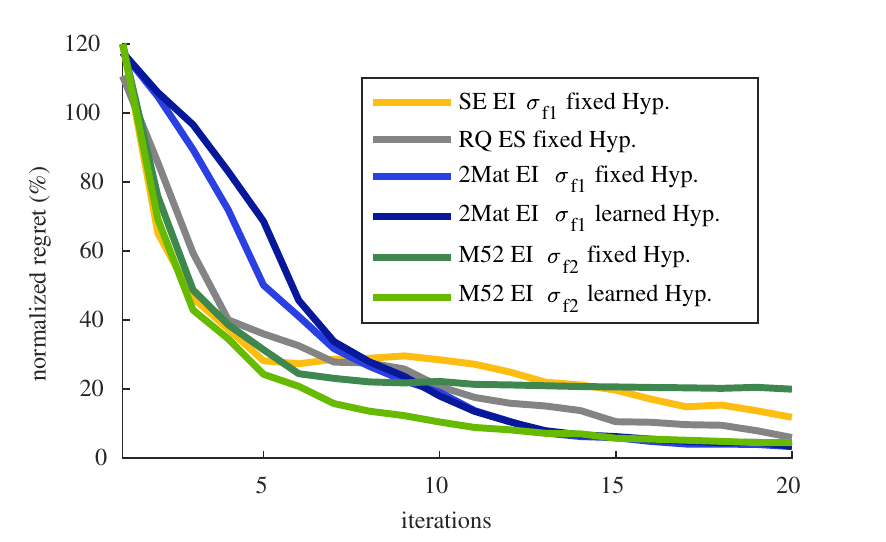}}
   \\
 \subfloat[Histogram of regret after 20 BO iterations\label{fig:comp_b}]{%
       \includegraphics[width=0.90\linewidth]{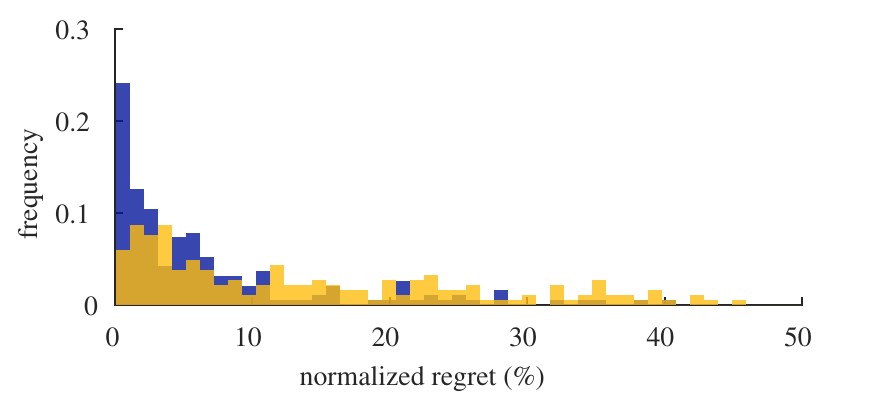}}
 \caption{(a) Comparison of the median over 200 run of the normalized regret for selected design choices at each iteration. (b) Histogram (normalized frequency) for \emph{SE EI $\sigma_{f1}$ fixed} in yellow and for \emph{2Mat  EI $\sigma_{f1}$ learned} in blue.
 }
 \label{fig:comp}
\end{figure}

The results are summarized in Table~\ref{tab:results}, which shows the median over the 200 runs of the normalized regret after 20 iterations.
The number in brackets represent the 95-percentile over the 200 runs of the normalized regret after 20 iterations; that is, the value that contains the best \SI{95}{\percent} of cases (in other words, \SI{95}{\percent} of the cases performed better than this value and only \SI{5}{\percent} worse).
This metric, which quantifies the robustness of the specific learning scheme, is arguably the most important in practice, as in real experiments, the gait shall usually only be learned once for each microrobot.
Numbers in bold represent the best performances achieved for the given design choices with and without hyperparameter optimization.
The settings plotted in Fig.~\ref{fig:comp} are filled grey.

One can see from the results that the SE kernel is (with one exception) outperformed by all other kernels.
The Matern-$3/2$ kernel and the sum of two Matern kernels of different length scales always perform reasonably well.

It can also be observed that MAP hyperparameter learning during BO does not lead to consistently better results.
Hyperparameter optimization can lead to inferior performance when directly compared to the same settings with fixed hyperparameters.
Optimizing the hyperparameter does however lead to better performance in cases where the set of fixed hyperparameters are performing relatively poorly.
Comparing design choices where the hyperparameter were seemingly  off (M52, EI,~$\sigma_{f2}$), learning the parameter helps significantly.
This is especially relevant for the performance in terms of the 95-percentile.
The best performing acquisition function was EI. We did, however, not test ES with hyperparameter optimization.

Figure~\ref{fig:comp_a} shows the comparison of a selection of design choices.
Examining the regret of the settings with the worst performance after 20 iterations, one can see it decreases cost faster and then flattens out after six to eight iterations.
We have observed that, after finding a local optimum, exploration stopped until the experimental budget was exhausted.
Hyperparameter optimization keeps the steep decrease in cost during the first iterations, but does not get stuck in a local optimum.
The histogram in Fig.~\ref{fig:comp_b} shows the distribution of cost for the blue and the yellow curves from Fig.~\ref{fig:comp_a}.
Not only is the median improved, but the spread is also reduced significantly.

The above comparison among different BO design choices allows us to evaluate the best setting to optimize the gait of a new, potentially different microrobot.
As discussed in Sec.~\ref{sec:introduction}, the learning scheme of choice has to be robust with respect to differences in microrobots and be highly data efficient, such that a good controller is found after 20 iterations.
As it can be observed in Table~\ref{tab:results}, the \emph{2Mat} kernel with EI and the optimistic signal variance~$\sigma_{f1}$ is expected to lead to substantial improvement in the locomotion performance, and to do so consistently.
As learning hyperparameters generally makes the learning scheme more robust, we selected the \emph{2Mat EI~$\sigma_{f1}$ learned Hyp.} configuration for gait learning.

%% file: sections/07_results.tex
\begin{figure}[thpb]
   \centering
   \includegraphics[width=0.90\linewidth]{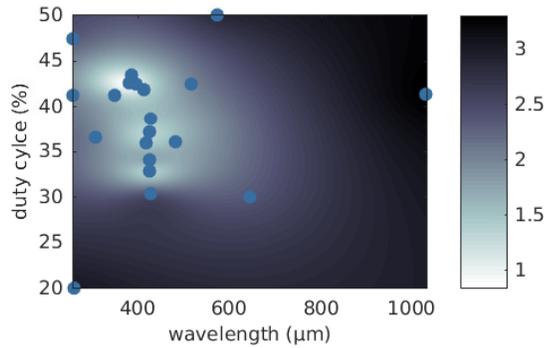}
   \caption{The posterior mean of the learned cost function in the parameter space. The blue dots show the 20 sample locations.}
   \label{fig:learning_experiment_cost}
\end{figure}

A new microrobot is used for the experimental validation of the developed learning control framework.
As each microrobot differs slightly, the optimal gait parameters are also expected to be different.
Without learning the specific cost function of the new microrobot the best known controller parameters are the ones identified in Sec.~\ref{sec:hyper}, namely $\lambda=\SI{645}{\micro\meter}$ and $\delta=\SI{30}{\percent}$.
Starting from this initial controller, we performed BO with an experimental budget of 20 samples.
The learned cost function over the parameter space is shown in Fig.~\ref{fig:learning_experiment_cost}.

\begin{figure}[thpb]
   \centering
   \includegraphics[]{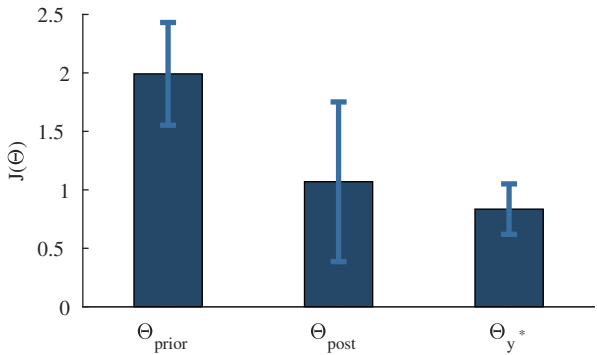}
   \caption{The cost for the initial, posterior and the best observed controller parameters and their standard deviation based on three evaluations at each parameters set.}
   \label{fig:learning_experiment}
\end{figure}

The initial performance was then compared to the performance at the parameter sets with lowest posterior mean ($\theta_{\text{post}}$) and with the lowest observed cost ($\theta_{y*}$).
The lowest-posterior-mean parameters were $\delta_{\text{post}}=\SI{43}{\percent}$ and $\lambda_{\text{post}}=\SI{376}{\micro\meter}$, whereas the lowest-observed-cost ones were $\delta_{y*}=\SI{43}{\percent}$ and $\lambda_{y*}=\SI{381}{\micro\meter}$.
Fig.~\ref{fig:learning_experiment} shows the obtained values of the cost function~$J(\theta)$.
Notably, we were able to reduce the cost by~\SI{46}{\percent} and~\SI{58}{\percent} in the lowest-posterior-mean and lowest-observed-cost cases, respectively.
Correspondigly, the locomotion performance, i.e.\ the average speed $V_m$, was increased by~\SI{92}{\percent} ($V_{m,\text{post}}=\SI{1.93}{\bls}$) and by~\SI{115}{\percent} ($V_{m,y*}=\SI{2.17}{\bls}$) with respect to the initial controller parameters obtained by previous experiments ($V_{m,\text{prior}}=\SI{1.01}{\bls}$).

%% file: sections/08_conclusion.tex
This paper is the first to investigate and demonstrate the potential of probabilistic learning of optimal gaits for sub-millimeter robots.
In particular, the proposed learning scheme based on BO with GPs proved to be highly data-efficient and robust against differences among microrobots.
These features are essential for light-driven soft microrobots, which can have a relatively high variability and a limited lifetime under continuous excitation.
The proposed BO learning approach is general and could be applied to find optimal controller parameters for other microrobotic systems.

The method we used to develop the learning scheme for gait optimization is based on the definition of a semi-synthetic benchmark function.
This approach allowed us to assess the robustness of different learning schemes against the inter-microrobot variability, and to select the scheme with the highest chance to provide substantial improvement in the locomotion performance of a slightly different microrobot.
This was verified in an actual microrobot experiment.
By using a different microrobot for the semi-synthetic benchmark and the validation experiment, we show that our method can handle differences between microrobots.

Learning control is an important step toward building a complete robotic system based on light-driven soft microrobots.
Still, there are many more interesting challenges ahead.
Future research directions are the design of problem-specific kernels, as well as the learning of more complex gaits in higher dimensional parameter spaces to take advantage of the microrobot's many DOFs~\cite{Palagi2016MARSS}.
This is relevant, for example, for planar locomotion using disk-shaped microrobots~\cite{Palagi2016Nat}.
In addition, extending BO to learn time-varying cost functions is also particularly interesting, as it will enable microrobots to adapt to degrading material properties and, more importantly, to changes in their environment.